\documentclass[square,comma,numbers]{article}

\usepackage[preprint]{neurips_2018}

\usepackage[utf8]{inputenc} 
\usepackage[T1]{fontenc}    
\usepackage{hyperref}       
\usepackage{url}            
\usepackage{booktabs}       
\usepackage{amsfonts}       
\usepackage{nicefrac}       
\usepackage{microtype}      
\usepackage{graphicx}	
\usepackage{subcaption}
\usepackage{amsmath}

\title{Neural Architecture Refinement: A Practical Way for Avoiding Overfitting in NAS}

\author{
Yang Jiang	\\
\texttt{jiangyang218@sina.com}	\\
\And
Cong Zhao	\\
\texttt{zhaocong.hk@gmail.com}	\\
\And
Zeyang Dou    \\
\texttt{douzeyang@qq.com}	\\
\And
Lei Pang		\\
\texttt{pangl01@pcl.ac.cn}	\\
}

\begin{document}

\maketitle

\begin{abstract}
Neural architecture search (NAS) is proposed to automate the architecture design process and attracts overwhelming interest from both academia and industry. However, it is confronted with overfitting issue due to the high-dimensional search space composed by operator selection and skip connection of each layer. This paper explores the architecture overfitting issue in depth based on the reinforcement learning-based NAS framework. We show that the policy gradient method has deep correlations with the cross entropy minimization. Based on this correlation, we further demonstrate that, though the reward of NAS is sparse, the policy gradient method implicitly assign the reward to all operations and skip connections based on the sampling frequency. However, due to the inaccurate reward estimation, curse of dimensionality problem and the hierachical structure of neural networks, reward charateristics for operators and skip connections have intrinsic differences, the assigned rewards for the skip connections are extremely noisy and inaccurate. To alleviate this problem, we propose a neural architecture refinement approach that working with an initial state-of-the-art network structure and only refining its operators. Extensive experiments have demonstrated that the proposed method can achieve fascinated results, including classification, face recognition etc. 
 
\end{abstract}

\section{Introduction}
\label{sec:intro}
 Due to the amazing feature representation power of deep neural networks (DNNs), DNNs have enabled remarkable progress on a variety of tasks, such as object detection\cite{Ren2015Faster,Liu2016SSD,Redmon2016You}, natural language processing\cite{Gehring2017Convolutional, Vaswani2017Attention, Devlin2018BERT}, speech recognition\cite{Graves2013Speech, Oord2016WaveNet}, face recognition\cite{deng2018arcface, fr2, fr3} etc. Nevertheless, the optimal network architecture designs vary dramatically for different tasks and hardware platforms, it is time-consuming and error-prone for machine learning experts to design different architectures. To address this problem, neural architecture search (NAS)  \cite{bello2017neural, zoph2018learning, zhong2018practical, pham2018efficient} has been proposed, it has attracted much attention in recent years.

NAS searches the optimal neural architecture in a search space. The element in the search space is a vector $\tau=[O_{1}, S_{1}, O_{2} \dots O_{n}, S_{n}]$, where $O_{m}$ is a one hot coding sub-vector representing the operation selection for node $m$, and $S_{m}$ is a binary sub-vector illustrating the skip connection between node $m$ and the previous nodes. The operators refer to either atomic operators such as conv3x3 or max-pooling, or fused operator composite such as conv3x3-relu-maxpolling. The skips are merely the connections between arbitrary two layers of operators. Because of the high dimensionality of the search space, searching an optimal architecture turns to be a difficult optimization problem. Numerous works therefore have been proposed to either reduce the search space or use a better optimization backend. For instance, \cite{zoph2018learning, liu2017hierarchical, liu2019auto} restricted
the diversity of the operator candidates and combined atomic operators into composite operators.
\cite{chen2018searching, zhong2018practical} proposed to search a structured cell rather than the whole architecture in a pre-defined
manner. As for optimization algorithm, \cite{liu2018progressive, van2018evolutionary, real2018regularized, liu2017hierarchical} employ the evolution algorithm, which is
computational expensive. Recently, reinforcement learning based methods (RL-based
methods)\cite{pham2018efficient, cai2018path, cai2018proxylessnas} and gradient-based method\cite{liu2018darts, cai2018proxylessnas, xie2018snas, Wu2018FBNet} are proposed, which requires
 much less computational resources and draw the attention of the field. Due to the large datasets and heavy computational cost of training, most of recent NAS can only search over a small sub dataset (which usually contains 1/10 samples), then transfer the found architecture to whole dataset. This strategy is known as “proxy task”.

Though the above discussed methods help to improve efficiencies of NAS, we argue that recent NAS frameworks suffer from the architecture overfitting problem, which prevent NAS from practical applications. Taking ENAS as an example, during the optimization of the architecture, the controller prefers to select architectures with dense skip connections, indicating that NAS easily falls into plateaus of over-complicated architectures, the search architectures would overfit the training task and perform poorly on unseen tasks.

In this paper, we explore the architecture overfitting problem in depth based on the RL-based NAS framework. We show that the policy gradient method has deep correlations with the cross entropy minimization. Based on this correlation, we further demonstrate that, though the reward of NAS is sparse, the policy gradient method implicitly assign the reward to all operations and skip connections. The assignment process is based on the final reward and the sampling frequency. Because these assigned rewards guide the optimization of the meta controller, it is important to build the reliable sampling strategy and estimate the final reward accurately. However, recent sampling strategies have strong bias, resulting in the inaccurate final reward estimation. Due to the limitations of computational resources, the architecture search process is often conducted on the proxy tasks, and the sampled architectures cannot be well trained. This approximation creates the search bias, because different architectures have different convergence rates. Given insufficient training steps, neural networks with shallow depths and intensive skip connections generally tend to achieve higher accuracy than those with deep depths and sparse skip connections. Besides, most of recent RL-based NAS framework jointly optimize operators and skip connections. However, as we will show, due to the curse of dimensionality of the skip connection space, the assigned reward suffers from the severe noise. This problem makes the loss surfaces of operators and skip connections have different characteristics, the optimization path of operators is smooth while that of skip connection is highly chaotic. This training gap would lead the model easily stuck into inferior local minima, raising the architecture overfitting issue. Moreover, due to the hierarchical structure of neural networks, the operator space and the skip connection space have different reward charateristics, using one controller is hard to model the joint search space.

Based on the above analysis, we propose a simple but effective method, which models the operator domain and the skip connection domain  separately and optimizes them alternatively, to alleviate the architecture overfitting problem. For optimizing the skip connection domain, because of the search bias and the chaotic optimization path of the skip connections domain, in principle we can optimize this domain using non-gradient methods (e.g., sampling), well evaluate all candidates and use the best one as the structure of the final model. However, this solution is computational expensive. In this paper, we propose a simple method which works with a initial network architecture and fixes its skip connections. In terms of the operator domain, due to the smoothness of its optimization path, we optimize this domain via policy gradient approach. We call this method Network Architecture Refinement (NAR). Because the skip connections is fixed during architecture search, NAR not only eliminates the search bias and curse of dimension problems, but also largely reduces the overfitting problem. Besides, the proposed method is easy and fast to converge.

We experimentally demonstrate that NAR can make many computer vision tasks achieve fascinated results, including classification, face recognition task (more results and tasks will present later on). In our experiment, NAR optimize the LResNet50E-IR\cite{deng2018arcface} and gain accuracy with 99.75\%, 97.61\% and 97.13\% compared to that with 99.68\%, 97.54\% and 96.86\% over LFW, CFP-FP, AgeDB-30 face recognition dataset with 24\% fewer FLOPS. 

\section{Related Work}
\label{sec:related}
For designing state-of-the-art architecture, NAS become the basic optimizer. Evolution-based NAS methods explore the network topological transformation by applying evolutional methods like crossover, mutation or recombination\cite{van2018evolutionary, real2018regularized,liu2017hierarchical}. Alternatively RL-based NAS and gradient-based NAS methods build a directed acyclic graph (DAG) and search a subgraph as the optimal architecture\cite{liu2018darts, cai2018proxylessnas, xie2018snas}. But current NAS methods are difficult to optimize because of the high dimensional search space.

Based on pre-defined search space that consists of the candidate operations and skip connections, NAS methods aim to find an optimal combination. The initial definition of operations is fine elements like ReLU activations, convolution layer, batch normalization, etc., which leads to huge search space so as to make NAS unpractical\cite{bello2017neural}. For reducing search space, \cite{zhong2018practical} and \cite{zoph2018learning} combine fine elements to build higher-level operations based on the hand-designed architectures like Conv-BN-ReLU, Depthwise convolution, etc. and obtain impressive performance. Meanwhile \cite{liu2019auto, chen2018searching} manually define some skip connection rules like recursive way etc. to constrain the connection degree of freedom. The attention of researcher turns from search space to topology\cite{liu2017hierarchical,xie2018snas}. Although the academic performance keep improved, the searched topologies may still suffer from overfitting as NAS still does not yet work in many practical tasks like face recognition.

\begin{table}
\centering
\begin{tabular}{|c|c|c|c|c|}
\hline
Architecture& Cifar&	LFW&	CFP-FP&	AgeDB30					\\
\hline
NAS-Arch&	\textbf{95.43}&	97.85&	88.97&	86.33\\
\hline
ResNet-20&			 94.15&		\textbf{98.36}&	\textbf{89.99}&	\textbf{88.28}\\	
\hline
\end{tabular}
\caption{Comparison of generalization between NAS-Arch and ResNet-20}
\label{table:1}
\end{table}

\section{Methodology}
\label{sec:overview}
\subsection{Definition of Architecture Overfitting}
Overfitting in machine learning refers to the problem that the well trained model cannot generalize well on the test data, because the model learns the non-robust patterns in the training data rather than the robust patterns of data. Architecture overfitting is defined in the same spirit, which means that the neural
architecture models the training task so well that its performance drops on the similar task. To investigate the architecture overfitting in NAS, we compare two neural architectures with similar
number of layers on two closely related tasks. The first architecture is a layer-wise searched architecture denoted as NAS-Arch\cite{pham2018efficient}, and the second is the handcrafted architecture
ResNet-20\cite{he2016deep}. The channels of ResNet-20 blocks is expanded to 64, 128 and 256 for fair comparison
with NAS-Arch. The NAS-Arch is searched on CIFAR10. Then NAS-Arch and ResNet-20 are trained and tested on CIFAR 10. Finally, we train NAS-Arch and ResNet-20 on face-emore \cite{guo2016ms, deng2018arcface} and test them on LFW\cite{Karam2015Quality}, CFP-FP \cite{Sengupta2016Frontal} and AgeDB30\cite{Moschoglou2017AgeDB, Deng2017Marginal}. Bear in mind that we use the same training strategy for NAS-Arch and ResNet-20. The result is
shown in table \ref{table:1}.
We see the NAS-Arch achieves better performance than that of ResNet-20 on Cifar, while ResNet-20 consistently outperforms NAS-Arch on three face recognition datasets. Since NAS-Arch has comparable parameters with ResNet-20, the performance gap doesn't come from the gaps of model capacities. This example demonstrates that NAS indeed suffers from the architecture overfitting problem.

\subsection{Understanding Architecture Overfitting}
We first show the relationships between the policy gradient method and the cross entropy minimization. Let $\tau^i = [O_{1}^{i}, S_{1}^{i}, O_{2}^{i} \dots O_{n}^{i}, S_{n}^{i}]$ be the $i$ $th$ sampled architecture, and $R^i$ is the corresponding reward. The policy gradient of reward expectation $ER$ is:
\begin{equation}
\nabla_{\theta} ER(\theta) = \frac{1}{N}\sum_{i}R^i\nabla_{\theta} log p(\tau^i|\theta),
\label{eq:1}
\end{equation}
where $\theta$ is parameters of the controller and $N$ is the number of sampled architectures. We factorize $p(\tau^i|\theta)$ as
\begin{equation}
p(\tau^i|\theta) = \prod_{j}p(O_{j}^{i}|Z_{1:j}^{i},\theta)p(S_{j}^{i}|Z_{1:j}^{i},O_{j}^{i},\theta)
\label{eq:2}
\end{equation}
where $Z_{1:j}$ represents the sampled operators and skip connections before node $j$. For notation simplicity, we use $p(O_{j})$ and $p(S_{j})$ to represent $p(O_{j}|Z_{1:j},\theta)$ and $p(S_{j}|Z_{1:j},O_{j},\theta)$ respectively. Substituting equation \ref{eq:2} into equation \ref{eq:1} yields
\begin{equation}
\nabla_{\theta} ER(\theta)=\frac{1}{N}\sum_{j}\sum_{i}R^i(\nabla_{\theta} log p(O_{j}^{i})+\nabla_{\theta} logp(S_{j}^{i})).
\label{eq:3}
\end{equation}
Equation \ref{eq:3} can be considered as the gradient of the following cross entropy loss
\begin{equation}
ER(\theta)=-\frac{1}{N}\sum_{j}\sum_{i}R^{i}\cdot CE(O_j^i,p(O_j^i))-\frac{1}{N}\sum_{j}\sum_{i}\sum_{t\leq j-2}R^{i}\cdot BCE(S_{j,t}^i,p(S_{j,t}^i)),
\label{eq:4}
\end{equation}
where $CE(x,y)$ and $BCE(x,y)$ are the cross entropy and the binary cross entropy with target distribution $x$ and predicted distribution $y$, and $S_{j,t}\in\{0,1\}$ is the binary number representing whether a skip connection exists between node $j$ and node $t$. We use $O_{j,k}\in\{0,1\}$ to denote whether the $j$ $th$ node samples the $k$ $th$ operator. Suppose we have $K$ operator candidates for each node, then the loss \ref{eq:4} can be rewritten as
\begin{equation}
ER(\theta)=\sum_{k=1}^{K}\sum_j(\frac{\sum_iO_{j,k}^iR^i}{N}) logp(O_{j,k})+\sum_{n=1}^{2}\sum_{j}\sum_{t\leq j-2}(\frac{\sum_i R^{i}\cdot \mathbf{S}_{j,t,n}^i}{N})logp(\mathbf{S}_{j,t,n}),
\label{eq:5}
\end{equation}
where $\mathbf{S}_{j,t,n} (n=1,2)$ is the one hot coding of $S_{j,t}$. The loss function \ref{eq:5} illustrates that the policy gradient implicitly assigns the final rewards to operators and skip connections based on their sampling frequency. Specifically, The reward for operator $O_{j,k}$ and $S_{j,t}$ are 
\begin{equation}
R_{O_{j,k}}=\frac{\sum_iO_{j,k}^iR^i}{N},
\label{eq:6}
\end{equation}
\begin{equation}
R_{S_{j,t}}= 
\begin{cases}
\frac{\sum_i R^{i}\cdot |{S}_{j,t}^i=1|}{N}& \text{if ${S}_{j,t}=1$}\\
\frac{\sum_i R^{i}\cdot |{S}_{j,t}^i=0|}{N}& \text{if ${S}_{j,t}=0$}
\end{cases}
\end{equation}
where
\begin{equation}
|{S}_{j,t}^i=1|=
\begin{cases}
1& \text{if ${S}_{j,t}=1$}\\
0& \text{if ${S}_{j,t}=0$}.
\end{cases}
\end{equation}

\begin{figure*}[t]
	\centering
	\begin{minipage}{1\linewidth}
	\centering
		\includegraphics[width=0.5\textwidth]{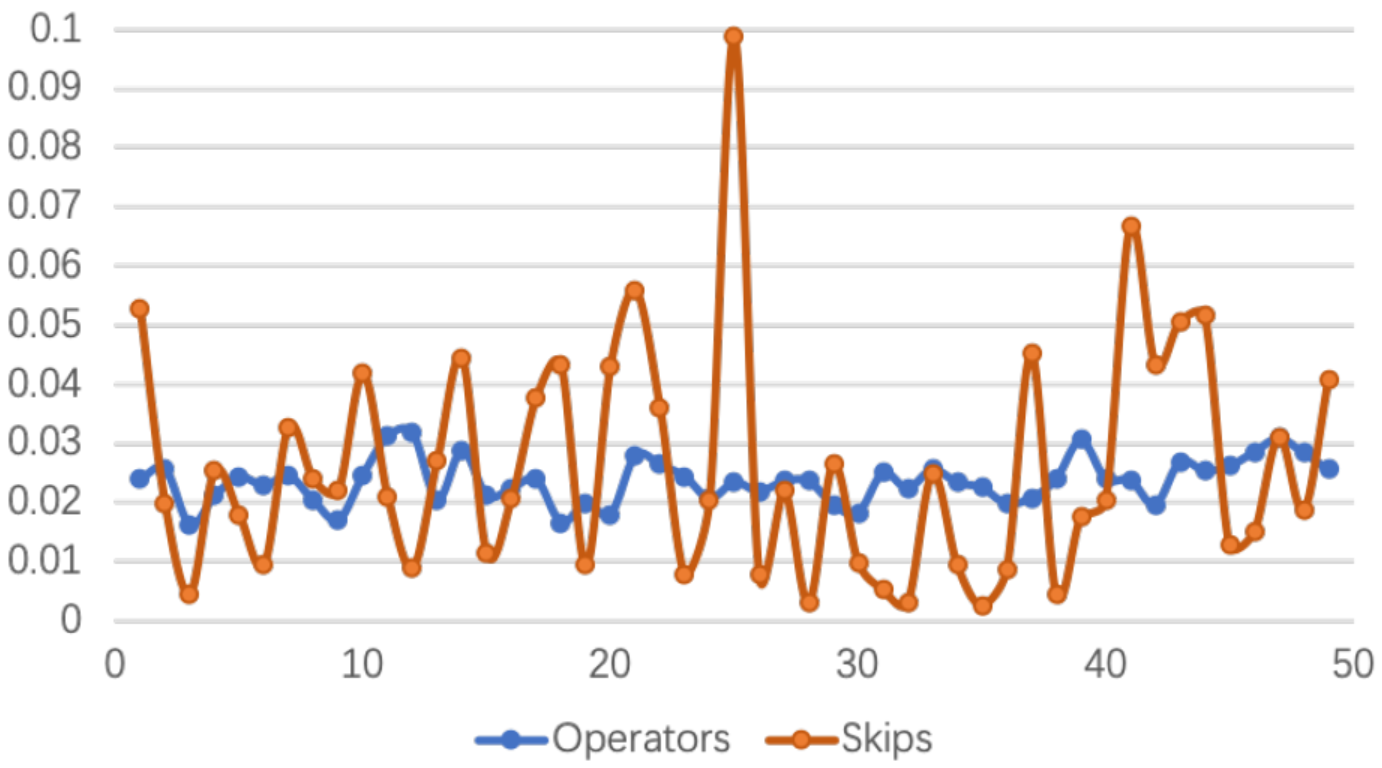}
	\end{minipage}

	\caption{Gradient magnitudes plot during architecture search. X- and Y-axis represent the training steps of meta-controller and the corresponding gradient magnitudes. The smoothness of two curves are totally different, indicating that operator and skip connections have different training difficulties.}
	\label{fig:1}
\end{figure*}

Obviously, the qualities of these assigned rewards heavily rely on the accurate estimation of $R^i$ and large $N$. However, recent sampling strategies have strong bias, resulting in the inaccurate $R^i$. Due to the limitations of computational resources, the architecture search process is often conducted on the proxy tasks, and the sampled architectures cannot be well trained. This approximation creates the search bias, because the networks with shallow depths and intensive skip connections usually have high validation accuracy at the early stage of training \cite{chen2019progressive, he2016deep}. 

Besides, $N$ is usually small because of the expensive training cost. However, the dimensionality of skip connection space is often much higher than that of operator space. For example, we assume that the architecture consists of 12 nodes, and each node has 6 operator candidates. The dimensionality of the operator space is $2\times 10^{9}$, while that of the skip connection space is $4\times10^{16}$. Sampling from this space suffers from the curse of dimensionality problem. Therefore, given small $N$, the sampling frequency is rather noisy and unstable. This problem brings optimization difficulty when the controller learns the optimal skip connections from the highly noisy samples. As shown in Figure \ref{fig:1}, the gradient of skip connections is oscillating while that of operators is far smoother. Therefore, the current reinforcement learning or continuous relaxation (e.g., DARTS) framework with one meta-controller is hard to model the joint space of operators and skips and the found architecture may overfit to the noisy skips in the training task and perform poorly on testing task.

\subsection{Neural Architecture Refinement}
\label{sec:method}
Based on the above analysis, we see that the search bias and the curse of dimensionality problem yield the optimization difficutlty of skip connection space. For the operator space, the search dimension of each node is the same, the assigned reward has the similar-level noise. However, for the skip connection space, the search dimension of the deep layers is much higher than that of shallow ones due to the hierachical structure of the network, the noise of the assigned reward become severe as the layer goes deep. Therefore, due to the different reward charateristics of the operator space and the skip connection space, using one LSTM meta-controller is hard to model the joint space.

From the perspective of feature representation learning, skip are usually expected for offering the different depth-level feature\cite{huang2017densely,huang2016deep}, and operator is mainly designed as the transformation-level function\cite{Sainath2015Deep,Krizhevsky2012ImageNet,zeiler2014visualizing}. Based on this fact, we argue that operator of each layer is related with the parameter complexity and skip connection accounts for the architecture complexity. Therefore, we define the operator and skip as parameter-overfitting related and architecture-overfitting related respectively, the operator and skip could be naturally considered as two domains. 

Based on above analysis, we split the operator and skip connections as two individual spaces, and alternatively optimize one space while fix another. Basically, the alternative search is searching the optima along two directions in the joint space of operator and skip. When searching alternatively, the reward in step $s-1$ for skip optimization is denotated as $ER^{s-1}$ and $ER^s$ represents the reward in step $s$ for operator optimization. Since our optimization strategy is optimizing one space by fixing another, we can reformulate (\ref{eq:5}) as

\begin{equation}
ER_{O}(\theta)=\sum_{k=1}^{K}\sum_j(\frac{\sum_iO_{j,k}^iR^i}{N}) logp(O_{j,k})+C_2,
\label{eq:9}
\end{equation}
\begin{equation}
ER_{S}(\theta)=C_1+\sum_{n=1}^{2}\sum_{j}\sum_{t\leq j-2}(\frac{\sum_i R^{i}\cdot \mathbf{S}_{j,t,n}^i}{N})logp(\mathbf{S}_{j,t,n}),
\label{eq:10}
\end{equation}
Where $C_1$ and $C_2$ are constants due to the fixed skips and operators. Suppose we begin the iteration from sampling the operation with fixed skips. We denote $O^{i}$ ($S^{i}$) as the sampled nodes (skips) at  $i$ th iteration. By using the policy gradient optimization framework, we can find  ($O^{i+1}$,$S^{i}$) such that $ER(O^{i+1},S^{i})\geq ER(O^{i},S^{i})$. Besides, the optimal nodes (skips) searched from the last iteration served as the initialization of the next iteration. Thus, by alternatively minimizing (\ref{eq:9}) and (\ref{eq:10}), we have
\begin{equation}
ER_{O}^{1}\leq ER_{S}^{1} \leq ER_{O}^{2} \leq ER_{S}^{2} \cdots \leq ER_{S}^{N}\leq maxER.
\label{eq:11}
\end{equation}
 The inequality (\ref{eq:11}) shows that $\{ER_{X}\}_{X \in \{O,S\}}$ is a monotone increasing bounded sequence, which indicates that we can find a Cauchy subsequence. This shows the convergence of the alternative optimization framework.

For optimizing the skip connection space, in principle we can optimize this domain using non-gradient methods (e.g., sampling), well evaluate all candidates and use the best one as the structure of the final model. However, this solution is computational expensive. Because the hand-crafted model structure has been verified its effectiveness, we propose a simple method that works with a initial network architecture and fixes its skip connections. In terms of the operator domain, due to the smoothness of its optimization path, we optimize this domain via policy gradient approach. Note that, due to the existence of $R^i$ in the operator reward (\ref{eq:6}), optimizing operators also have the search bias which might lead to less diversity of operations. To address this problem, we pre-train the graph randomly for few epochs before starting to search. The pretraining would help to alleviate the search bias of the early stage of training. In the experiments, we denote the method using pre-train with suffix ``PreTrain". We choose RL-based NAS as our operator optimizer. Following ENAS\cite{pham2018efficient}, we use LSTM\cite{hochreiter1997long} as meta-controller that generate combination of operations and train the shared parameters of child models. The proposed method largely reduces the search space. Besides, since the skip connections and the model depth are fixed, we don't have the curse of dimensionality problem and the search bias is largely reduced.

\subsection{The Definition of Search Space}
\label{subsec:definition}
We define the rules that how to build search space. First, given the task that need to be applied by NAS, we take the state-of-the-art architecture that based on current research as our base model and do not modify the skip connections. Non-modified skips can be seen as higher level kind of high level combined operations, like the combination of skips and operations. Then, the search space composed by $operator$ and $skip$ is reduced to $operator$ only.

Following the works\cite{pham2018efficient} we take their operations as our operator candidates, $conv 3*3, conv 5*5, depthwise 3*3, depthwise 5*5, max 3*3, avg 3*3$. Note that we treat the $max 3*3$ and $avg 3*3$ as non-parameterized convolution-like operator. Given one architecture, we prefer to choose the candidate operators with no more parameters than the architecture operators to avoid parameter-overfitting.

For example, we take face recognition as our task in this paper and follow the rules, we take $conv 3*3, depthwise 3*3, max 3*3, avg 3*3$ as candidate operations since the face model only use $conv 3*3$.

\section{Experiment}
\label{sec:exp}
We evaluate our proposed method on two kinds of tasks, face recognition and image classification.
The initial architectures used in these two tasks have distinctive characteristics, such as number of
parameters, number of layers etc. From the experiment, we want to demonstrate that our method
have wide adaptability for both small and large architectures. All of our architecture refinements are
conducted on one NVIDIA 1080Ti and implemented with PyTorch.

\subsection{Face Recognition}

For face recognition, ArcFace\cite{deng2018arcface} model is chosen as our baseline. We use NAR with and without pre-train to refine LResNet50E-IR \cite{deng2018arcface} and name the refined models as LResNet50E-IR-R and LResNet50E-IR-RP. The
refinement is conducted on a proxy task and the refined architectures are trained over the whole
dataset. The generalizability of the refined architecture is verified by training on two different tasks
with performance reserving. 

\subsubsection{Dataset and Proxy Task}

We choose face emore V1\cite{guo2016ms,Deng2017Marginal} and V2\cite{guo2016ms,deng2018arcface} as our training data, which consist 384,846 images with 85,164 identities and 5,822,653 images with 85,742 identities respectively. Distribution of face emore V1 and V2 are both non-uniform.
Since face emore datasets are large, we build our proxy task for efficiency. We only randomly collect 160k images from face emore V2 as our proxy data. The proxy dataset contains 4k identities and each identity consists of 40 images. We random select 36 images as the training data for each identity and the rest images are used for validation. We use SGDR\cite{loshchilov2016sgdr} optimizer with $T_{mult}$=2 ,$T_{0}$=10, max learning rate = 0.1 and min learning rate = 0.0001. The training stops after 150 epochs. As for data augmentation, we follow the same strategy as \cite{deng2018arcface}.

We use LFW, CFP and AgeDB dataset as our test data:
\begin{itemize}
\item \textbf{LFW}\cite{Karam2015Quality}: LFW dataset contains 5749 different identities with 12,233 web-collected images. These images vary in pose, expression and illuminations. We use 6,000 face pairs by the standard protocol of unrestricted with labeled outside data.
\item \textbf{CFP}\cite{Sengupta2016Frontal}: CFP contains 500 subjects, each of which consists of 10 frontal and 4 profile images. There are 10 folders with 350 same-person pairs and 350 different-person pairs in each of the evaluation protocol and the protocol includes frontal-frontal (FF) and frontal-profile (FP) face verification. We only choose CFP-FP, which is the challenging subset of CFP.
\item \textbf{AgeDB}\cite{Moschoglou2017AgeDB,Deng2017Marginal}: AgeDB contains 440 subjects with 12,240 images of varied pose, expression, illuminations, and age. The average age is 49 years with the minimum of 3 and maximum of 101.Test data of AgeDB divided into four groups with different year gaps, that is 5 years, 10 years, 20 years and 30 years. There are 10 spit images in each group, and each split includes 300 positive examples and 300 negative examples. We evaluate our model on AgeDB, that is the challenging subset.
\end{itemize}

\subsubsection{Architecture Analysis}

The refined architecture vectors are shown in Table \ref{tab:fr_arch}, where we denotate $conv3\times3$; $depthwise3\times3$; $maxpool3\times3$; $avgpool3\times3$ as 0,1,2,3
respectively. Obviously, the vector of baseline is full of zero. In addition, architecture with more
diverse operators are generated with pretrain strategy. Table \ref{tab:fr_result} shows the computational load comparisons of all models, including the number of parameters and the FLOPS. We see that the refined models have less Flops and parameters than their baselines.

\begin{table}
\centering
\begin{tabular}{|c|c|}
\hline
Model							&				Architecture					\\	\hline
LResNet50E-IR					&	[0,0,0,0,0,0,0,0,0,0,0,0,0,0,0,0,0,0,0,0,0,0,0,0]		\\	\hline
LResNet50E-IR-R		&	[0,1,2,0,0,0,0,0,1,0,0,0,0,0,0,0,0,0,0,0,0,0,1,3]		\\	\hline
LResNet50E-IR-RP	&	[0,0,2,0,0,0,1,0,0,0,0,0,0,0,0,1,1,0,2,2,1,0,1,0]		\\	\hline
\end{tabular}
\caption{Baseline and searched architectures}
\label{tab:fr_arch}
\end{table}

\subsubsection{Performance Analysis}
We train the refined architectures on face emore V2 and test them on LFW, CFP, and AgeDB respectively. For comparison, we train LResNet50E-IR in the same environment as our benchmark. The hyper-parameter configuration is the same as ArcFace. We use SGDR\cite{loshchilov2016sgdr} training strategy with $T_{mult}$=2 ,$T_{0}$=5, max learning rate = 0.1 and min learning rate = 0.0001, and train for total 35 epochs. 

\begin{table}
\centering
\begin{tabular}{|c|c|c|c|c|c|}
\hline
Methods		&	LFW(\%)		&	CFP-FP(\%)	&	AgeDB-30(\%)		&	Parameter Number	&	FLOPS	\\	\hline
LResNet50E-IR&	99.68		&	97.54		&	96.86			&	$4.35\times10^7$	&	$6.32\times10^9$	\\	\hline
LResNet50E-IR-R		&	99.73		&	97.59		&	\textbf{97.32}		&	$3.40\times10^7$	&	$5.31\times10^9$	\\	\hline
LResNet50E-IR-RP		&	99.75		&	97.61			&	\textbf{97.13}		&	$3.38\times10^7$	&	$4.69\times10^9$	\\	\hline
\end{tabular}
\caption{LResNet50E-IR, LResNet50E-IR-R and LResNet50E-IR-RP trained over Face-emore V2}
\label{tab:fr_result}
\end{table}

From Table\ref{tab:fr_result}, we can see that the LResNet50E-IR-RP trained on face emore V2 gain 99.75\%, 97.61\% and 97.13\% compared to 99.68\%, 97.54\% and 96.86\% over LFW, CFP-FP, AgeDB-30 face recognition dataset with 24\% fewer FLOPS. LResNet50E-IR-R obtains the similar results. The refined models
maintain the regular performance over variation in pose and illumination. Furthermore, they are robust
to the frontal-profile and wide age range of the same person. It is important to note that NAR improves the baseline on all test datasets with much less FLOPS, indicating that NAR eases the architecture-overfitting issue in handcrafted architecture.

\subsubsection{Generalization}
To further verify the generalization, we directly train the refined architectures on face emore V1. Note that the architectures are searched on proxy task of a different dataset, face emore V2. The results are shown in Table\ref{tab:trans_fr_result}. We can see that Both LResNet50E-IR-R and LResNet50E-IR-RP that trained over V1 can also achieve better performance over three datasets, which indicates that the searched architectures have better capability of generalization.

\begin{table}
\centering
\begin{tabular}{|c|c|c|c|c|c|}
\hline
Methods		&	LFW(\%)		&	CFP-FP(\%)	&	AgeDB-30(\%)		&	Parameter Number	&	FLOPS	\\	\hline
LResNet50E-IR&	99.55		&	93.64		&	95.51			&	$4.35\times10^7$	&	$6.32\times10^9$	\\	\hline
LResNet50E-IR-R		&	99.55		&	93.76		&	95.55			&	$3.40\times10^7$	&	$5.31\times10^9$	\\	\hline
LResNet50E-IR-RP		&	99.56		&	\textbf{94.42}	&	95.60			&	$3.38\times10^7$	&	$4.69\times10^9$	\\
\hline
\end{tabular}
\caption{LResNet50E-IR, LResNet50E-IR-R and LResNet50E-IR-RP trained over Face-emore V1}
\label{tab:trans_fr_result}
\end{table}

\subsection{Classification Task}

For the classification task, we select ResNet-18\cite{he2016deep} as our baseline and refine it on CIFAR-10\cite{krizhevsky2009learning}. We evaluate the searched model on CIFAR-10 and then transfer the refined model to ImageNet\cite{russakovsky2015imagenet}.

\subsubsection{Architecture Analysis}

Our experiment follows the search strategy\cite{he2016deep}. We search the residual block in two ways: 1) treat the whole residual block as one elements in search space, 2) treat each CONV-BN-RELU as one elements in search space. There are 8 depth layers for the first strategy and 16 for the second. We call the first architecture as ResNet-18-8 and the second as ResNet-18-16. The searched architectures are shown in Table\ref{tab:cls_arch}. Because the search space of ResNet-18-8 is relatively small, there is just one operator being replaced. When we search using the second strategy, four operators are changed. ResNet-18-8 and ResNet-18-16 have 11\% and 20\% less FLOPS than baseline, even though ResNet-18 is already relatively small.

We see that NAR works better when we optimize a large and deep model. Literally, Large and deep model is much easier to overfit, which also demonstrate that NAR is practical for easing  overfitting issues.

\begin{table}
\centering
\begin{tabular}{|c|c|c|c|}
\hline
Model		&	Architecture		&	Parameter Number		&	FLOPS			\\	\hline
ResNet-18	&	[0,0,0,0,0,0,0,0]	&	$1.17\times10^7$		&	$1.76\times10^9$	\\	\hline
ResNet-18-8	&	[1,0,0,0,0,0,0,0]		&	$1.16\times10^7$		&	$1.57\times10^9$	\\	\hline
ResNet-18-16	&	[1,0,0,2,0,0,0,0,1,0,0,0,0,0,0,2]		&	$0.92\times10^7$		&	$1.41\times10^9$	\\
\hline
\end{tabular}
\caption{Architecture of ResNet-18 and ResNet-18-8, ResNet-18-16}
\label{tab:cls_arch}
\end{table}

\subsubsection{Results for classification}

We evaluate ResNet-18-8 and ResNet-18-16 over CIFAR-10 and then transfer them to ImageNet for generalization test. The results are shown in Table\ref{tab:trans_cls_result}. The refined models obtain slightly better performance on CIFAR-10 and comparable results on ImageNet but with fewer FLOPS.

The results of the classification task also demonstrate that NAR can not only improve the model but also has good capability of generalization.

\begin{table}[t]
\centering
\begin{tabular}{|c|c|c|c|}
\hline
Model		&	CIFAR-10(\%)		&	ImageNet Top-1(\%)		&	ImageNet Top-5(\%)	\\	\hline
ResNet-18	&	93.46			&	69.39					&	89.09				\\	\hline
ResNet-18-8	&	93.72			&	69.42					&	89.10				\\	\hline
ResNet-18-16	&	93.71				&	69.18					&	88.81				\\
\hline
\end{tabular}
\caption{Results of ResNet-18 and ResNet-18-8, ResNet-18-16 over CIFAR and ImageNet}
\label{tab:trans_cls_result}
\end{table}

\section{Conclusion}
\label{sec:conclusion}
In this paper, we explored the architecture overfitting problem in depth based on the reinforcement learning NAS framework. We showed that the policy gradient method has deep correlations with the cross entropy minimization. Though the reward of NAS is sparse, the policy gradient method implicitly assigns the reward to all operations and skip connections based on the reward and the sampling frequency. However, due to the inaccurate reward estimation, curse of dimensionality problem and the hierachical structure of neural networks, reward charateristics for operators and skip connections have intrinsic differences, the assigned rewards for the skip connections are extremely noisy and inaccurate. Based on this analysis, we proposed a simple method, named as neural architecture refinement (NAR), to refine neural architectures. NAR only focuses on optimizing operator space while keep the skip connections fixed. NAR largely reduces FLOPs and parameters in the refined architecture, estimates the reward more accurate and alleviates the curse of dimensionality problem. Several transfer learning studies were conducted to further verify the generalization ability. Finally, despite that NAR achieved consistent performance in different computer vision tasks, skip connection, as the basic elemental elements of neural architecture, could be better modeled and sampled. Since operators and skips have different contributions to architecture design, searching the optimal architecture in an alternative way worth further investigation.

\bibliographystyle{unsrt}
\bibliography{refs}

\end{document}